# On the Efficiency of the Neuro-Fuzzy Classifier for User Knowledge Modeling Systems

Ehsan Jeihaninejad[1] and Azam Rabiee[2,*]

[1,2] Young Researchers and Elite Club, Dolatabad Branch, Islamic Azad University, Isfahan, Iran
[1]Jeihani.ehsan@gmail.com
[2] azrabiee@gmail.com

**Abstract.** User knowledge modeling systems are used as the most effective technology for grabbing new user's attention. Moreover, the quality of service (QOS) is increased by these intelligent services. This paper proposes two user knowledge classifiers based on artificial neural networks used as one of the influential parts of knowledge modeling systems. We employed multi-layer perceptron (MLP) and adaptive neural fuzzy inference system (ANFIS) as the classifiers. Moreover, we used real data contains the user's degree of study time, repetition number, their performance in exam, as well as the learning percentage, as our classifier's inputs. Compared with well-known methods like KNN and Bayesian classifiers used in other research with the same data sets, our experiments present better performance. Although, the number of samples in the train set is not large enough, the performance of the neuro-fuzzy classifier in the test set is 98.6% which is the best result in comparison with others. However, the comparison of MLP toward the ANFIS results presents performance reduction, although the MLP performance is more efficient than other methods like Bayesian and KNN. As our goal is evaluating and reporting the efficiency of a neuro-fuzzy classifier for user knowledge modeling systems, we utilized many different evaluation metrics such as Receiver Operating Characteristic and the Area Under its Curve, Total Accuracy, and Kappa statistics.

## 1    Introduction

An automatic User modeling system is adapted to present user requirements and conditions using collected information from users or customers for designing a dynamic or static model. In the world wide web, these systems are known as the most effective technology for recognizing users. In addition, they can help to make "the right service at the right time" idea possible by their adaptability.

Making the system more customized and adaptable to user requirements is the idea behind the user modelling. This could be possible by having a model or system's internal representation of the users and their information. Since the system definition and applicability depend on information classes, the type of the provided information plays the important role in designing the model. User's characteristics, such as interests, skills, age, and private information (e.g., name, weight) are some data used by such systems.

Regarding the type of the data, there are four categories of user modeling systems, illustrated in Figure 1. As showed in the figure, some models are defined as static model since they use data that are hardly changeable (e.g., name, belief) or unchangeable like birthday [1]. In contrast, the second category is called dynamic model (i.e, user knowledge model) as they are designed by changeable data like skills and knowledge [2-3].

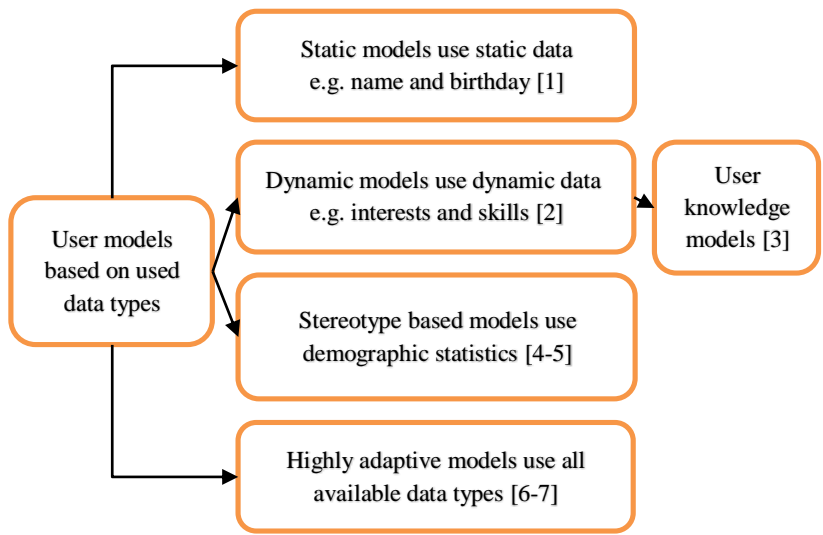

**Fig. 1.** Various user models based on the used data type.

Model belonged to The third category are designed by statistical data like demographic statistics [4-5] in which personal information rarely take into account. Finally, the last category includes the models that are designed by all provided dynamic and static data. as a result, the model adaptability is remarkable, although the cost of designing such models is high [6-7].

in this paper we focus on classifying users based on their dynamic information such as educational, learning, and skills. Concisely, our main goal is classifying users using their knowledge level in a way that the classification method can be used as an important part of a user knowledge modeling system. Such systems are usually designed based on a special domain of entities, so that there are always a strong relationship between the considered domain and the designed user knowledge model [8].

One of the main parts of the user knowledge modeling system is its classifier algorithms. Some of the applications of the classifiers in the user knowledge modeling systems are 1) determining and recommending research domain of the user; 2) the goal of their study; and 3) specifying student knowledge level. In this paper, we demonstrate that the classification algorithm based on neuro-fuzzy networks can significantly classify users' knowledge level.

The rest of the paper is organized as follows. First, in Section 2, we give an overview of the related studies on users modeling systems. Designed model and used dataset's specification are discussed in Section 3. The experimental results followed by comparing with other methods' results will be presented in 4th Section. Finally, in Section 5, we summarize the paper.

## 2    Related Studies

Web-based applications such as social networks, benefit from the user models. To improve the web services, the presented content should be more adaptable to user's interest. A solution to keep the user safe from being confused of various recommendations is to use users' rating behaviors methods. Yin and his colleagues [9] believe that such methods pave the way for emerging new applications such as personalized recommendation, computational advertising, and information filtering. Regarding the study, there are two factors which have a great influence on users' rating behaviors modeling: 1) the external influence, and 2) the internal interest.

The attention of the general public as an external factor, is more challenging than the internal factor, the intrinsic interests of the user, because the former is dynamic but the latter is more stable. The aforementioned research has presented DTCAM which is the dynamic model of TCAM and is more complete as user dynamic interests take into account.

Nowadays, one of the major concerns is designing review rating prediction systems. These systems predict the user's rating which is a number from 1 to 5. The rating is performed about a text considering the words of its content. However, Tang and his colleagues [10] discussed that, if the focus in predicting is only based on the used words, the system prediction cannot be reliable enough. They believe that a text is interpreted by author's characteristics and personality along with the concepts of the used words, meaning that using the same word in different sentences which is written by two authors could have different meaning. In the study they reported that, the user model, which is called user information, should be combined with the user review rating prediction system to make the result reliable enough. A branch of the user modeling is user knowledge modeling which is utilized to create intelligent systems like real time student knowledge assistant [2].

Artificial intelligence (AI) plays a vital role in the field of user modeling [11-12]. One of the AI applications in this filed, is designing classifiers for user knowledge modeling systems. But dataset's samples shortage has always been a big challenge for training the classifiers so that this shortage can lead to performance reduction in most classifying techniques which are based on weight adjustment like artificial neural networks.

Kahraman and his colleagues [8] in 2013, classified users' knowledge levels using Bayesian classifier and K-nearest neighbor (KNN). In the above mentioned paper, a genetic algorithm based technique called intuitive knowledge classifier (IKC) is proposed for classifying the user knowledge level which outperformed Bayesian and KNN performances. in this paper we use IKC results to show the performance of the proposed model.

## 3    Proposed model

In this research, we investigated the effect of using adaptive network-based fuzzy inference system (ANFIS) algorithm on the user knowledge level classification. ANFIS was first proposed  by Jang in 1993 [13] to overcome one of the most challenging issues of fuzzy inference systems (FIS), known as expert's knowledge acquisition. This acquisition causes many problems such as dependency of the system to the expert, since knowledge is naturally dynamic. As a result, updating the system needs the expert to be always present.  The other problem is the challenge of expert's knowledge acquisition for the knowledge engineer. Moreover, Automatic adjustment of FIS parameters like membership functions, is another challenging issue. ANFIS overcome these problems by automatically analyzing the gathered data from considered domain. Classification, event prediction, and pattern recognition are some problems that can be solved by ANFIS algorithm.

   We designed a 4-layers ANFIS network for classifying user's knowledge level which is demonstrated by Figure 2.

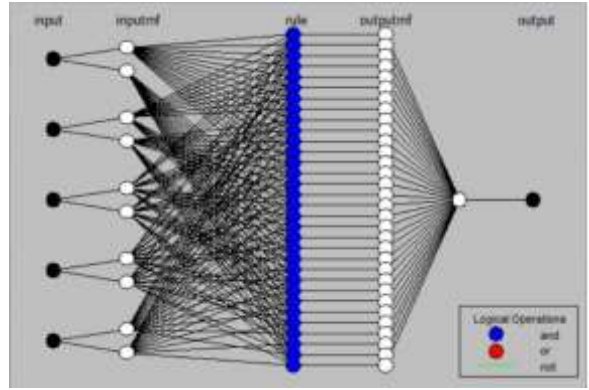

**Fig. 2.** the structure of designed ANFIS network.

The first layer presents that each input belongs to fuzzy set with its specified membership degree.  Actually this layer adjusts the inputs' membership functions in an automatic manner. In our experiments, we used Gaussian, and triangular membership functions.  Formula 1 is the used Gaussian function and Figure 3, depicts the triangular function.

$$O_i = \mu(x) = \frac{1}{1 + \left(\frac{(x - c_i)}{a_i}\right)^{2b_i}} \tag{1}$$

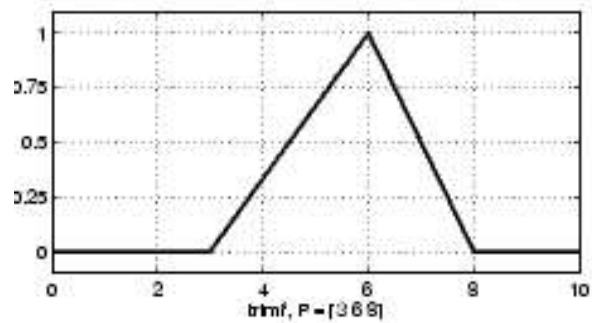

**Fig. 3.** the triangular function.

The second layer is served as Firing Strength generator which is the result of applying the AND and OR operators on the second layer outputs. This value specifies that how much a specific rule can be true for different input values. the designed network has a normalizer part which is the third layer. Each of this layer output is calculated by the fraction of each previous layer output and summation of all previous layer outputs. This calculation is based on Formula 2.

$$O_i = \overline{W_i} = \frac{w_i}{\sum_i w_i} \tag{2}$$

Finally, in the fourth layer, the output signals of the previous layer are added together to generate the final result. Note that, for the computations we used MATLAB and neuro-fuzzy system is based on Sugeno's fuzzy inference method. It is worthy of note that the ANFIS output is decimal so to use the output as the classification result, an ordinary threshold function is used.

### 3.1 User knowledge modeling dataset

In this research, we used Kahraman's real user knowledge modeling dataset [8]. The dataset contains 6 attributes shown in Table 1 and totally have 403 samples. The attribute information of the dataset is as follows.

- STG: the degree of study time.
- SCG: the degree of repetition number.
- PEG: the user performance in exam.
- STR: the degree of study time for prerequisite objects.
- LPR: learning percentage of prerequisite objects.

The first three above mentioned attributes are related to the learning objects and the rest are about the prerequisite objects which are served as the user knowledge classifier inputs. The attributes' data are digits between 0 and 1 which are mapped to -1 and 1 because of our classifying strategy. The classifier output is UNS which refers to user knowledge level and takes one of alphabetic values, very low, low, middle, and high. Distribution of these four classes are demonstrated in Table 2.

**Table 1.** the used dataset specifications.

| | Input | Output |
|---|---|---|
| Attribute | STG, SCG, STR, LPR, PEG | UNS |
| Data type | Integer | Symbolic |
| Data values | -1 and 1 | Very low, low, middle, high |

**Table 2.** the distribution of dataset classes.

| Class | The # of samples |
|---|---|
| Very low | 50 |
| low | 129 |
| Middle | 122 |
| High | 130 |

## 4    Experimental results

In the experiments, we divided the dataset into training and testing data containing 80 and 20 present of dataset samples respectively. After training the ANFIS classifier, we used the test data to check how accurately it can classify user knowledge levels. To show the performance of our approach we compared the ANFIS classifier performance with the performance of methods proposed in [8].

In the mentioned research, Kahraman and his colleagues proposed an intuitive user knowledge classifier which is a hybrid classifier based on genetic algorithms. To compare IKC performance, Kahraman used K-nearest neighbor and Bayesian knowledge classifier results. The abbreviations used in Table 3, EU, MA, and MI, are various distance metrics applied for IKC method and are Euclidean, Manhattan, and Minkowski respectively.

To compare more, we designed a feed forward backpropagation neural network knowledge classifier in which the number of neurons and layers were determined in a trial and error manner. It is worth mentioning that the cross-validation technique was considered in the experiments and the performances shown in table 3 are the maximum classifiers accuracy.

### 4.1    Evaluation Metrics

**Total Accuracy.**
To compare the results of IKC and proposed neural based models, we utilize one of the well-known single value metric named Total Accuracy by which it can be shown that how accurately a model classifies the classes. The accuracy is calculated using dividing the sum of True Positive (TP) and True Negative (TN) by sum of TP, TN, False Positive (FP), and False Negative (FN).

The experimental results shown in Table 3 demonstrates that the ANFIS classifier can classify user knowledge levels accurately and shows better performance toward IKC method. In addition, the performance of 2-layer feed forward ANN classifier overcomes Bayesian and KNN performances. Although The ANN performance was not better than the IKC (in case of accuracy) but it shows a comparable result.

**Table 3.** the results of proposed classifier and the comparison of designed method with other classifiers proposed in [8]. IKC is a hybrid method based on genetic algorithms in which MI, EU, and MU are the most popular distance metrics named Minkowski, Euclidean, and Manhattan. Moreover, in this table, B, MWCS, and CAP stand for Bayesian, The Mean of Wrong Classified Samples, and Classifying Accuracy in Percentage respectively.

| Classifier method | ANN based proposed classifiers | | IKC | | | K-nearest neighbor | | | B |
|---|---|---|---|---|---|---|---|---|---|
| algorithm | ANFIS | ANN | EU | MA | MI | BJ | MA | MI | |
| MWCS | 2 | 3.5 | 3 | 3 | 5 | 26.2 | 21.7 | 30.5 | 38 |
| CAP | 98.62 | 97.24 | 97.9 | 97.9 | 96.5 | 81.9 | 85 | 79 | 73.8 |

**Kappa Statistic and AUC.**
In this section we compare the results of ANN and ANFIS models using the well-known Area Under the Curve (AUC) of Receiver Operating Characteristic (ROC) and Kappa metrics. Kappa which is calculated by Formula 3 is another single value evaluation metric by which the accuracy of the system to a random system's accuracy (hypothetical expected probability of agreement) is compared.

$$Kappa = \frac{TotalAccuracy + RandomAccuracy}{1 - RandomAccuracy} \qquad (3)$$

In which *RandomAccuracy* is calculated using Formula 4.

$$RandomAccuracy = \frac{(TN+FP)*(TN+FN)+(TP+FN)*(TP+FP)}{(total)*(total)} \qquad (4)$$

In which total presents the sum of TP, TN, FP, and FN.

For evaluating Kappa, we utilized the One Against All (OAA) approach to obtain confusion matrixes values. Tables 4 and 5 and Figures 4 and 5 demonstrate the results of OAA approach along with the estimated values for Kappa and AUC metrics. As shown by table 5, we used Gaussian2 and Triangular-Shaped (trimf) Matlab built-in Membership Functions (MF) to evaluate the ANFIS model's output. Moreover, reported by table 4, transfer functions used for ANN model are Hyperbolic Tangent Sigmoid (tansig) and Log-Sigmoid (logsig). Illustrated tables and figures of AUC and Kappa evaluations report that, ANN classifier show better performance toward ANFIS when using OAA technique.

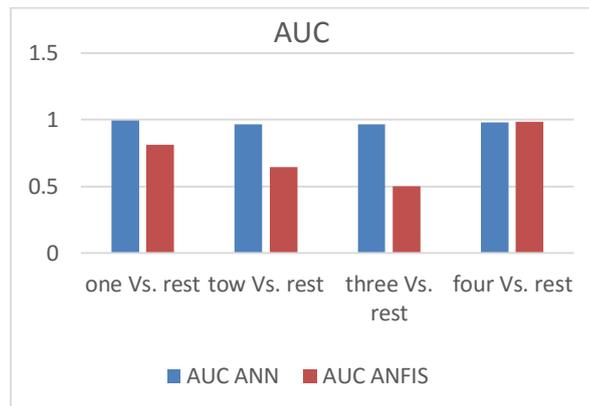

**Fig. 4.** comparison of AUC values when using ANN and ANFIS classifiers

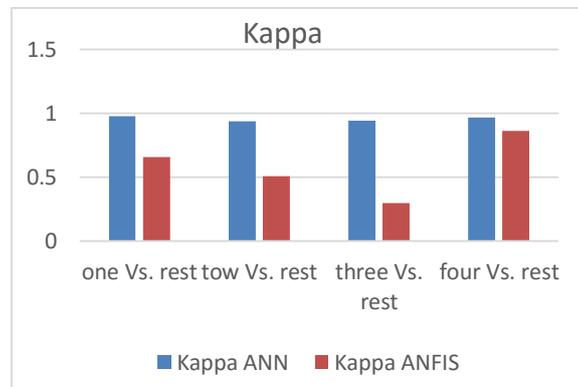

**Fig. 5.** comparison of Kappa values when using ANN and ANFIS classifiers

## 5    Conclusion

User modeling systems are one of the most crucial parts of web based applications. Regarding the data type used for designing, these systems differ from each other and have different applications. Systems which are designed based on user dynamic information (e.g. skills, educations) are called user knowledge models. In this paper an important part of these systems (classifier) which is based on neuro-fuzzy networks was proposed. In other words, the goal of our study was classifying users' knowledge level using ANFIS classifier and analyzing the efficiency of using neuro-fuzzy based classifiers toward other approaches proposed in the literature. in this research, we used a real dataset whose samples are obtained by students' interactions in the web.

Experimental results show that ANFIS can classify users' knowledge level more accurately than the other methods. We also used a feed-forward backpropagation neural network classifier which leaded to performance reduction compared with ANFIS but its performance was better than KNN and Bayesian approaches.

**Table 4.** the calculated values for Kappa and AUC when using ANN classifier.

| classes | transfer function | TPR | FPR | TNR | FNR | total accuracy | Random accuracy | Kappa | AUC |
|---------|---------|-----|-----|-----|-----|------|------|-------|-----|
| one Vs. rest | tansig | 1 | 0.008 | 0.99 | 0 | 0.99 | 0.701 | 0.976 | 0.9958 |
| | logsig | 1 | 0 | 1 | 0 | 1 | 0.705 | 1 | 1 |
| tow Vs. rest | tansig | 0.97 | 0.04 | 0.95 | 0.02 | 0.972 | 0.566 | 0.936 | 0.9682 |
| | logsig | 0.97 | 0.02 | 0.97 | 0.02 | 0.979 | 0.564 | 0.952 | 0.979 |
| three Vs. rest | tansig | 0.99 | 0.05 | 0.94 | 0.009 | 0.979 | 0.644 | 0.941 | 0.9661 |
| | logsig | 0.98 | 0.11 | 0.88 | 0.01 | 0.95 | 0.648 | 0.882 | 0.9322 |
| four Vs. rest | tansig | 0.99 | 0.02 | 0.97 | 0.009 | 0.986 | 0.606 | 0.964 | 0.9825 |
| | logsig | 0.98 | 0.02 | 0.97 | 0.018 | 0.979 | 0.603 | 0.947 | 0.9777 |

**Table 5.** the calculated values for Kappa and AUC when using ANFIS classifier.

| classes | MF | TPR | FPR | TNR | FNR | total accuracy | random accuracy | Kappa ANFIS | AUC |
|---------|-----|-----|-----|-----|-----|------|------|------|-----|
| one Vs. rest | gauss2mf | 0.53 | 0 | 1 | 0.46 | 0.91 | 0.75 | 0.65 | 0.769 |
| | trimf | 0.65 | 0.02 | 0.97 | 0.34 | 0.91 | 0.73 | 0.69 | 0.814 |
| tow Vs. rest | gauss2mf | 0.67 | 0.08 | 0.91 | 0.32 | 0.75 | 0.49 | 0.50 | 0.794 |
| | trimf | 0.76 | 0.47 | 0.52 | 0.23 | 0.68 | 0.56 | 0.28 | 0.644 |
| three Vs. rest | gauss2mf | 0.82 | 0.52 | 0.47 | 0.17 | 0.74 | 0.63 | 0.29 | 0.649 |
| | trimf | 0.85 | 0.85 | 0.14 | 0.14 | 0.68 | 0.68 | 0.003 | 0.501 |
| four Vs. rest | gauss2mf | 0.95 | 0.07 | 0.92 | 0.04 | 0.94 | 0.60 | 0.86 | 0.938 |
| | trimf | 0.97 | 0 | 1 | 0.02 | 0.97 | 0.59 | 0.94 | 0.985 |

# 6    References

1.  J. Hothi and W. Hall, "An evaluation of adapted hypermedia techniques using static user modelling," *2nd WS Adapt. Hypertext Hypermedia, USA*, 1998.

2.  J. P. Rowe and J. C. Lester, "Modeling User Knowledge with Dynamic Bayesian Networks in Interactive Narrative Environments," *Sixth AAAI Conf. Artif. Intell. Interact. Digit. Entertain.*, pp. 57–62, 2010.

3.  A. Hawalah and M. Fasli, "A multi-agent system using ontological user profiles for dynamic user modelling," *Proc. - 2011 IEEE/WIC/ACM Int. Conf. Web Intell. WI 2011*, vol. 1, pp. 430–437, 2011.

4.  R. Alnanih, O. Ormandjieva, and T. Radhakrishnan, "Context-based user stereotype model for mobile user interfaces in health care applications," *Procedia Comput. Sci.*, vol. 19, pp. 1020–1027, 2013.

5.  K. Chrysafiadi and M. Virvou, "Student modeling approaches: A literature review for the last decade," *Expert Syst. Appl.*, vol. 40, no. 11, pp. 4715–4729, 2013.

6.  F. Carmagnola, F. Cena, and C. Gena, "User model interoperability: A survey," *User Model. User-adapt. Interact.*, vol. 21, no. 3, pp. 285–331, 2011.

7.  T. Lavie and J. Meyer, "Benefits and costs of adaptive user interfaces," *Int. J. Hum. Comput. Stud.*, vol. 68, no. 8, pp. 508–524, 2010.

8.  H. T. Kahraman, S. Sagiroglu, and I. Colak, "The development of intuitive knowledge classifier and the modeling of domain dependent data," *Knowledge-Based Syst.*, vol. 37, pp. 283–295, 2013.

9.  H. Yin, B. Cui, L. Chen, Z. Hu, and X. Zhou, "Dynamic User Modeling in Social Media Systems," *ACM Trans. Inf. Syst.*, vol. 33, no. 3, pp. 1–44, 2015.

10. D. Tang, B. Qin, T. Liu, and Y. Yang, "User modeling with neural network for review rating prediction," *IJCAI Int. Jt. Conf. Artif. Intell.*, vol. 2015–Janua, no. Ijcai, pp. 1340–1346, 2015.

11. S. Zhang, F. Boukamp, and J. Teizer, "Ontology-based semantic modeling of construction safety knowledge: Towards automated safety planning for job hazard analysis (JHA)," *Autom. Constr.*, vol. 52, pp. 29–41, 2015.

12. G. Stalidis, D. Karapistolis, and A. Vafeiadis, "Marketing Decision Support Using Artificial Intelligence and Knowledge Modeling: Application to Tourist Destination Management," *Procedia - Soc. Behav. Sci.*, vol. 175, pp. 106–113, 2015.

13. J. S. R. Jang, "ANFIS: Adaptive-Network-Based Fuzzy Inference System," *IEEE Trans. Syst. Man Cybern.*, vol. 23, no. 3, pp. 665–685, 1993.